\documentclass[onecolumn]{article}

\usepackage{amsmath,amssymb,amsfonts,amsthm,enumerate,multirow,mathtools}
\usepackage{color}
\usepackage{siunitx}
\usepackage{parskip}
\usepackage[hidelinks]{hyperref}
\usepackage{placeins} 
\usepackage{booktabs}
\usepackage{algorithm,algpseudocode}
\usepackage[affil-it]{authblk}
\usepackage[margin=2.5cm]{geometry}
\usepackage[round]{natbib}
\usepackage{subcaption}
\usepackage{enumitem}

\newcommand{\parsection}[1]{\vspace{2mm}\noindent\textbf{#1}~ }

\title{Deep Energy-Based NARX Models}

\author[1]{Johannes N. Hendriks}
\author[2]{Fredrik K. Gustafsson}
\author[3]{Ant\^onio H. Ribeiro}
\author[1]{Adrian G. Wills}
\author[2]{Thomas B. Sch\"on}

\affil[1]{School of Engineering, The University of Newcastle, Callaghan NSW 2308, Australia (email: \{johannes.hendriks,adrian.wills\}@newcastle.edu.au)}
\affil[2]{Department of Information Technology, Uppsala University, Uppsala, Sweden (e-mail: \{fredrik.gustafsson,thomas.schon\}@it.uu.se)}
\affil[3]{Department of Computer Science, Federal University of Minas Gerais, 31270-901 Belo Horizonte, Brazil (antoniohorta@dcc.ufmg.br}

\begin{document}
\maketitle


\begin{abstract}
This paper is directed towards the problem of learning nonlinear ARX models based on system input--output data. In particular, our interest is in learning a conditional distribution of the current output based on a finite window of past inputs and outputs. To achieve this, we consider the use of so-called energy-based models, which have been developed in allied fields for learning unknown distributions based on data. This energy-based model relies on a general function to describe the distribution, and here we consider a deep neural network for this purpose. The primary benefit of this approach is that it is capable of learning both simple and highly complex noise models, which we demonstrate on simulated and experimental data.
\end{abstract}



\section{Introduction}\label{sec:introduction}

This paper considers the problem of learning a model for dynamic systems based on observed system input--output data. This problem has a long and fruitful history within the system identification, statistics and machine learning communities and there are many different ways to approach it. For example, a regularly employed approach is to first define a suitable parameterized model structure based on knowledge of the system. Then we learn, adapt, infer or estimate the parameters based on the available evidence in the data. To decide between different parameters, and ultimately provide the best values, the user is required to choose a performance criterion such as the maximum-likelihood (ML) or prediction-error criteria.

It is important to note that both the model structure and estimation method involve assumptions about uncertainty, be they explicit or implicit. That is, the probability distribution that represents this uncertainty is assumed. For example, it is not uncommon that users explicitly assume additive white Gaussian noise as a way of modelling measured output uncertainty. Further, it can be argued that this same assumption is implicit in mean-squared-error estimation. More generally, in many practical situations, it is difficult to simply justify these assumptions from the available prior system knowledge or even from the data.

This paper details a means for addressing this difficulty by allowing the distribution itself to be modelled using a highly flexible function that is learned from the available data. The primary benefit of this approach is that it can easily adapt to both highly complex distributions and also less complicated ones such as a unimodal Gaussian. The inspiration for this approach comes from the allied field of machine learning where so-called energy-based models (EBMs), typically combined with deep neural networks (DNN's), are employed for modelling unknown distributions with great success \citep{du2019implicit, Grathwohl2020Your, gustafsson2020energy}.


To make these ideas concrete, this paper will concentrate on the class of nonlinear-autoregressive-exogenous-input (NARX) dynamic models \citep{ljung1999system}. In particular, it will be assumed that the current system output $y_t$ is related to past outputs $y_{t-1},\ldots,y_{t-D_{y}}$, and past inputs $u_{t-1},\ldots,u_{t-{D_u}}$; where $D_y$ is the maximum output delay and $D_u$ is the maximum input delay. Our particular interest here is in providing a conditional distribution of $y_t$ given the past data window. That is, we are concerned with describing
\begin{equation}
  \label{eq:probabilistic_form}
  y_t | x_t \sim p(y_t | x_t),
\end{equation}
where $x_t$ contains the past data window:
\begin{align}
  \label{eq:x_def}
  x_t = \{y_{t-1},\ldots,y_{t-D_y},u_{t-1},\ldots,u_{t-D_u}\}.
\end{align}
Unfortunately, it is not immediately obvious how to choose this distribution so that it explains measured system data. One way to address this difficulty is to assume a functional form for this distribution that relies on some unknown parameters $\theta$, which we denote as $p_\theta(y_t | x_t)$. The idea then is to estimate these parameter values based on the available evidence in the data. This raises at least two questions; how should we parameterise this distribution, and, how should we learn from the data?

Regarding the first problem of parameterisation, a traditional approach for NARX models is to first formulate an output equation form
\begin{equation}
  \label{eq:functional_form}
  y_t = f_\theta(x_t) + e_t,
\end{equation}
where $f_\theta$ is a function that is traditionally linear in the parameters $\theta$, but is otherwise quite a general function of the past data $x_t$. The added term $e_t$ is a random variable that characterises the error between the function output $f_\theta(x_t)$ and the measured output $y_t$, and, its distribution may also depend on $\theta$. Therefore, by construction, the conditional distribution of interest, $p_\theta(y_t | x_t)$, will depend on the assumed choice of distribution for $e_t$.

Regarding the second problem of learning from the data, again a traditional approach is to formulate and solve the associated ML problem \citep{ljung1999system}. By way of a concrete example, assuming that $y_t \in \mathbb{R}$ and the noise $e_t$ is i.i.d. Gaussian with zero mean and variance $\sigma^2$, then the ML solution for $\theta$ coincides with
\begin{align}
  \label{eq:mse_crit}
  \widehat{\theta} = \arg \min_\theta \sum_{t=1}^T \| y_t - f_\theta(x_t) \|^2.
\end{align}
Therefore, an estimate of the desired conditional distribution $p(y_t | x_t)$ is given by
\begin{align}
  \label{eq:5}
  y_t | x_t \sim \mathcal{N}(f_{\hat{\theta}}(x_t), \sigma^2).
\end{align}
More complex distributions for $e_t$ can also be accommodated within the ML framework, but this requires the user to choose a suitable distributional family. In many practical situations, it is not obvious how to select this family based on prior system knowledge.

This paper aims to address this difficulty by providing a highly flexible class of distributions that are adapted to each new problem based on the available system data. In particular, $p(y_t | x_t)$ will be modelled with the conditional EBM $p_\theta(y_t | x_t) = e^{g_\theta(y_t, x_t)} / \int e^{g_\theta(\gamma, x_t)}\,\mathrm{d}\gamma$, where the scalar function $g_\theta$ is represented by a DNN with associated parameters $\theta$. This energy-based approach puts very few restricting assumptions on the true distribution $p(y_t|x_t)$, enabling it to be learned directly from data. 

\parsection{Contributions}
The main contribution of this paper is an energy-based model capable of learning $p(y_t|x_t)$ for dynamic systems. We evaluate the new construction on both simulated and experimental data, demonstrating its benefits compared to more traditional NARX models. This paper thus illustrates the utility of EBMs and their potential within system identification.

\section{Related Work}

During the last decade, there has been a surge of interest in DNN models
and these models have been used to obtain
state-of-the-art solutions for many applications, including computer
vision, speech recognition and natural language processing
\citep{lecun_deep_2015}. While the use of neural networks in
system identification problems has a long history~\citep{narendra_identification_1990, chen_nonlinear_1990}, the
success of the method in neighbouring areas has brought a new wave of
interest within the system identification
community~\citep{ljung_deep_2020}, with recent papers leveraging
acquired knowledge and being inspired by successful ideas from recent
DNN applications. Examples of deep-learning-inspired
ideas applied in system identification include; convolutional network
layers \citep{andersson_deep_2019}, encoder-decoder
structure~\citep{gedon_deep_2020} and recurrent neural networks and
its extensions~\citep{gedon_deep_2020,ljung_deep_2020}.

EBMs have been extensively studied by the machine learning community \citep{lecun2006tutorial, teh2003energy, osadchy2005synergistic}. They are usually employed for unsupervised learning applications, and have in recent years become particularly popular for generative modelling within computer vision \citep{nijkamp2019learning, du2019implicit, Grathwohl2020Your}. In comparison, the application of EBMs to supervised learning problems is not a very well-studied topic, but their effectiveness has been demonstrated for both classification \citep{ma2018noise} and regression \citep{gustafsson2020energy}. Most closely related to our proposed approach is the very recent work on employing conditional EBM's for regression \citep{gustafsson2020energy, danelljan2020probabilistic, gustafsson2020train}, achieving state-of-the-art performance on tasks such as object detection and tracking.

\section{Energy-Based NARX Models} 
\label{sec:deep_energy_based_models}
Inspired by \citet{gustafsson2020energy}, we model the distribution $p(y_t | x_t)$ with the conditional EBM
\begin{equation}\label{eq:conditional_EBM}
  p_\theta(y_t | x_t) = \frac{e^{g_\theta(y_t, x_t)}}{\int e^{g_\theta(\gamma, x_t)}\,\mathrm{d}\gamma},
\end{equation}
where $g_\theta$ is a DNN that maps any pair $(y_t, x_t)$ directly to a scalar $g_\theta(y_t, x_t) \in \mathbb{R}$. 

Here, $p_\theta(y_t | x_t)$ is directly specified via the DNN $g_\theta$, which provides a highly flexible
class of functions. This enables $p_\theta(y_t | x_t)$ to model a wide range of
distributions, including heavy-tailed, asymmetric or multimodal ones. Related to this, we note that the DNN output value
$g_\theta(y_t, x_t) \in \mathbb{R}$ is proportional to the logarithm
of the distribution $p_\theta(y_t | x_t) $, not to the output $y_t$ itself. This has
implications for how the model may be used, which will be discussed in
Section~\ref{sub:prediction_using_the_deep_ebm} below.



Evaluating the denominator $Z(x_t) = \int g_\theta(y_t | x_t)\,\mathrm{d}y_t$ in~(\ref{eq:conditional_EBM}) presents a challenge since this
integral is analytically intractable in general. For the case when $y_t$ is low-dimensional, the integral may be evaluated using standard
quadrature methods. In the more general case, we advocate the use of
Monte Carlo methods for solving this integral (see \citet{gustafsson2020energy} for details on this approach).


Since the EBM (\ref{eq:conditional_EBM}) relies on a nonlinear
combination of previous data $x_t$, we will refer to this as an energy-based NARX (EB-NARX)
model.
Next, we first provide more details on the structure of the DNN $g_\theta$ in Section~\ref{sub:neural_network_structure}. We then describe how to learn the unknown DNN parameters $\theta$ based on a set of training data $\mathcal{D} = \{y_t, x_t\}_{t=1}^T$, in Section~\ref{sub:training}. Finally, we discuss how the model
can be used for prediction, in Section~\ref{sub:prediction_using_the_deep_ebm}.


\subsection{Neural Network Structure} 
\label{sub:neural_network_structure}

\begin{figure}[htb]
    \centering
    \includegraphics[width=0.6\linewidth]{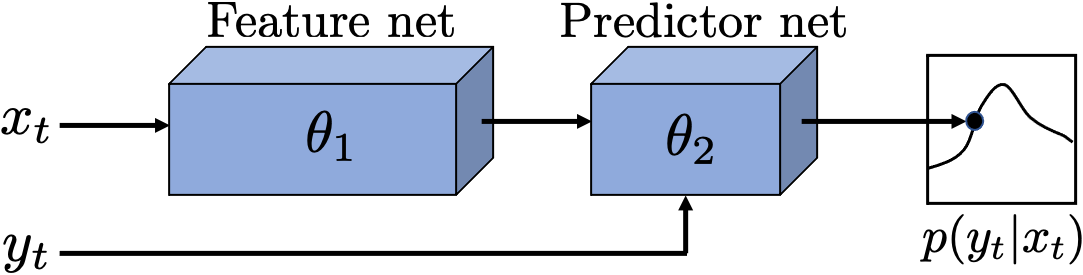}
    \caption{Structure of the deep EB-NARX model used.}
    \label{fig:ebm_structure}
\end{figure}

The DNN $g_\theta$ is composed of two smaller neural
networks; a feature net and a predictor net parametrised by $\theta_1$
and $\theta_2$, respectively. The feature net takes $x_t$ as input and produces a feature vector. This feature is then combined with $y_t$ and fed as input to the predictor net, which finally outputs the unnormalised log density $g_\theta(y_t, x_t) \in \mathbb{R}$ of \eqref{eq:conditional_EBM}. See Figure~\ref{fig:ebm_structure} for an illustration. This structure has the benefit that when making predictions the feature net only needs to be evaluated once, after which the predictor net can be evaluated for a range of $y_t$ values.


\subsection{Training the Neural Network}
\label{sub:training}
Presented with the data $\mathcal{D} = \{y_t, x_t\}_{t=1}^T$ and the DNN $g_\theta(y_t, x_t)$, it is tempting to consider the
ML problem as a means for learning the parameters $\theta$. Towards
this, we can express the joint likelihood, under the assumption of
independence, as
\begin{align}
  \label{eq:2}
  p_\theta(y_{1:T} | u_{1:T}) &= p_\theta(y_T | y_{1:T-1},
                                   u_{1:T})\  p_\theta(y_{1:T-1} | u_{1:T}),
\end{align}
where we have used conditional probability to arrive at the expression
on the right. Noting the assumed temporal and causal nature of the
NARX model, then repeated application of conditional probability
delivers
\begin{align}
  p_\theta(y_{1:T} \mid u_{1:T}) = \prod_{t=1}^T p_\theta(y_t \mid
                                   x_t)
  = \prod_{t=1}^T\frac{e^{g_\theta(y_t,
    x_t)}}{\int e^{g_\theta(\gamma, x_t)}\,\mathrm{d}\gamma}. \label{eq:prod_gtheta}
\end{align}
Therefore, the ML estimate for $\theta$ coincides with
\begin{align}
  \widehat{\theta}  &= \arg \max_\theta  p_\theta(y_{1:T} \mid
                     u_{1:T}),\\
                    &= \arg \min_\theta  -\ln p_\theta(y_{1:T} \mid
                     u_{1:T}),\\
                    &= \arg \min_\theta \sum_{t=1}^T \left (
                      - g_\theta(y_t, x_t) + \ln
                      \int e^{g_\theta(\gamma , x_t)}\,\mathrm{d}\gamma
                      \right ),
\end{align}
where the second equality relies on logarithm being a monotonic
operator, which implies that the solutions coincide. The third
equality is simply the negative logarithm applied to
\eqref{eq:prod_gtheta}. 
This ML problem is not immediately soluble, due to the analytically intractable integral. Numerical integration can however be employed to obtain an approximate solution, as shown in \citep{gustafsson2020energy}.


Alternative approaches for fitting a distribution $p_\theta(y_t | x_t)$ to observed data $\{y_t, x_t\}_{t=1}^T$ can also be applied to learn the parameters $\theta$. This was studied in detail for conditional EBMs by \cite{gustafsson2020train}, recommending noise contrastive estimation (NCE) \citep{gutmann2010noise} over ML. We thus employ NCE and learn $\theta$ by minimizing the cost function $L(\theta)\!=\!-\frac{1}{T} \sum_{t = 1}^{T} L_t(\theta)$,
\begin{equation}
    \label{eq:nce}
    L_t(\theta) = \ln \frac{\exp\left(g_\theta(y_t^{(0)},x_t)\!-\!\ln q(y_t^{(0)} | y_t)\right)}{\sum_{m=0}^M\exp\left(g_\theta(y_t^{(m)},x_t)\!-\!\ln q(y_t^{(m)} | y_t)\right)},
\end{equation}
where $y_t^{(0)}\triangleq y_t$, and $\{y_t^{(m)}\}_{m=1}^M$ are $M$ noise samples drawn from $q(y | y_t)$. This noise distribution is a mixture of $K$ Gaussians centered at $y_t$,
\begin{equation}
    q(y | y_t) = \frac{1}{K}\sum_{k=1}^K\mathcal{N}(y | y_t, \sigma_k^2 I).
\end{equation}
Since \eqref{eq:nce} can be interpreted as the cross-entropy loss for a classification problem with $M+1$ classes, NCE intuitively entails learning to discriminate between the output $y_t$ and sampled noise $\{y_t^{(m)}\}_{m=1}^M$.




\subsection{Prediction using the Deep EBM} 
\label{sub:prediction_using_the_deep_ebm}
Rather than giving a point prediction, the proposed deep EB-NARX model
predicts $g_\theta(y_t, x_t) \propto \ln p_\theta(y_t | x_t)$.
There are two ways in which this prediction could be used: if the
uncertainty of the prediction is important then we can evaluate
$p_\theta(y_t | x_t)$; alternatively, if we only require a point
estimate then we could choose the maximum a posterior (MAP) estimate.

The MAP estimate, $\hat{y}_{t}$, can be found by solving
\begin{equation}
    \hat{y}_t = \arg \max_{y_t} p_\theta(y_t | x_t) = \arg \max_{y_t} g_\theta(y_t, x_t).
\end{equation}
Since there is no guarantee that $p_\theta(y_t | x_t)$ is unimodal, it
was found practical to evaluate $g_\theta(y_t, x_t)$ for a spread of
values and then refine the best of these using gradient ascent, $y_t \gets y_t + \lambda \nabla_{y_t} g_{\theta}(y_t, x_t)$.

An estimate of $p_\theta(y_t | x_t)$ can be determined by evaluating
\eqref{eq:conditional_EBM} for a range of feasible values of $y_t$,
where the denominator can be determined by numerical integration, such
as Monte Carlo integration.



\section{Examples} 
\label{sec:examples}
This section provides several examples which illustrate the utility of
the EB-NARX model when applied to data from dynamic systems. These
examples include both simulated linear and non-linear data, as well as
real data from the CE8 coupled electric drives nonlinear data set
\citep{wigren2017coupled}. For the linear examples, qualitative
comparisons are made between the estimated and true distributions. For
the non-linear examples, qualitative comparisons are made between a
fully connected network (FCN) and EB-NARX estimates of the conditional
distributions.  

While simple, FCN's obtain highly competitive results in nonlinear system identification benchmarks, even when compared with more sophisticated approaches, such as convolutional and recurrent neural networks, see the benchmarks in~\citet{andersson_deep_2019}. The FCN models are estimated in the functional form \eqref{eq:functional_form}, nonetheless the conversion to a probabilistic form \eqref{eq:probabilistic_form} is straightforward: we use the implicit assumption of Gaussian noise (which is made when minimizing the least square cost function), where the mean is the output of the model and the variance is the sample variance.


Quantitative comparison between the EB-NARX model
estimates and the true values are given using the mean squared error
(MSE) based on the MAP value from the predicted conditional
distribution.

Python code for these examples is available at \url{https://github.com/jnh277/ebm_arx}.

\subsection{Pedagogical Example} 
\label{sub:pedagogical_scalar_example}
\begin{figure}[!htb]
    \centering
    \subcaptionbox{Gaussian\label{subfig:scalar_gauss}}{
    \includegraphics[width=0.3\linewidth]{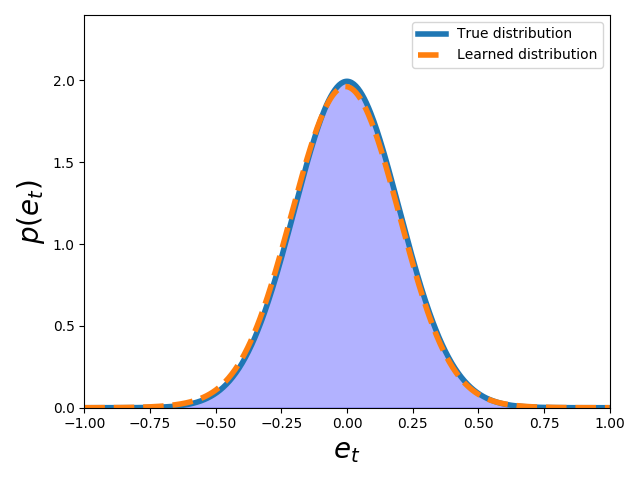}}
\subcaptionbox{Bimodal Gaussian\label{subfig:scalar_bimodal}}{
    \includegraphics[width=0.3\linewidth]{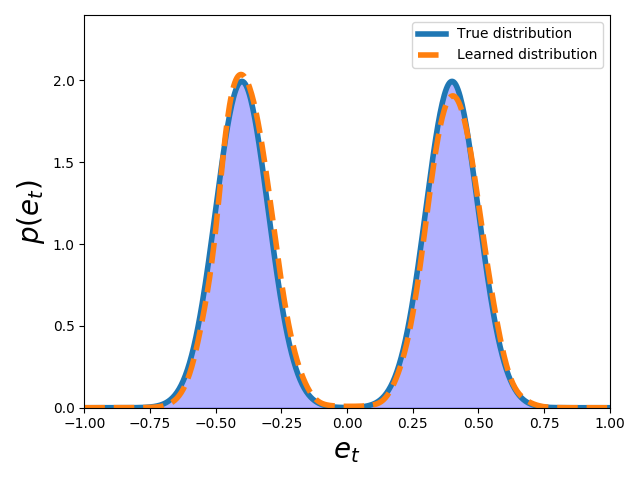}} \\
\subcaptionbox{Cauchy\label{subfig:scalar_cauchy}}{
    \includegraphics[width=0.3\linewidth]{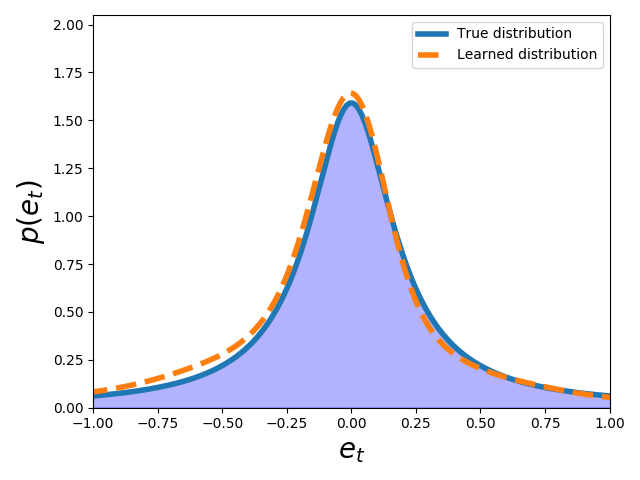}}
\subcaptionbox{Dependent variance \\ Gaussian\label{subfig:dep_variance}}{
    \includegraphics[width=0.3\linewidth]{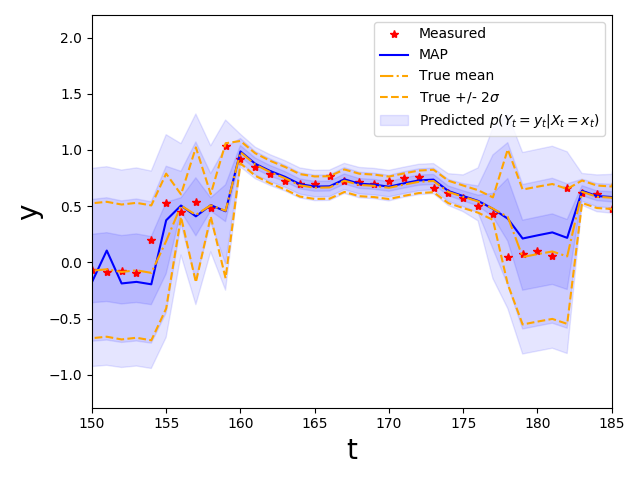}}
    \caption{Pedagogical example of learning different distributions using a deep EB-NARX model from data generated using a simple AR model \eqref{eq:ar_model}. }
    \label{fig:ar}
\end{figure}

First, the ability of the EB-NARX model to learn different distributions is illustrated.
To do this, the method is applied to data generated using a simple autoregressive (AR) model with different distributions for the noise;
\begin{equation}
    \label{eq:ar_model}
    y_t = 0.95 y_{t-1} + e_t.
\end{equation}
Four different distributions for the noise $e_t$ are considered:
\begin{enumerate}[label=\alph*)]
    \item zero-mean Gaussian, $e_t \sim \mathcal{N}(0,0.2^2)$,
    \item bimodal Gaussian, {\small $e_t \sim 0.5\mathcal{N}(0.4,0.1^2) + 0.5\mathcal{N}(-0.4,0.1^2)$},
    \item zero-mean Cauchy, $e_t \sim \mathcal{C}(0, 0.2^2)$,
    \item Gaussian with variance dependent on the systems state,
    \begin{equation}
        e_t ~\sim \begin{cases}\mathcal{N}(0,0.3^2)\qquad  &\text{if} \quad |y_{t-1}| < 0.5 \\
        \mathcal{N}(0,0.05^2) \quad &\text{otherwise}.
        \end{cases}
    \end{equation}
\end{enumerate}
The learned distributions are shown in Figure~\ref{fig:ar}. While Gaussian noise is often a fair assumption, the utility of a more flexible noise model is made apparent by considering that measurement outliers can be modelled by Student's T or Cauchy distributions. Moreover, in Section~\ref{sub:real_data_coupled_electric_drives} the real data gives rise to distribution that is conditional on $x_t$ and in some cases bimodal.

\subsection{Linear ARX} 
\label{sub:linear_arx}
\begin{figure}[!htb]
    \centering
    \subcaptionbox{Sequence\label{subfig:arx_sequence}}{
    \includegraphics[width=0.3\linewidth]{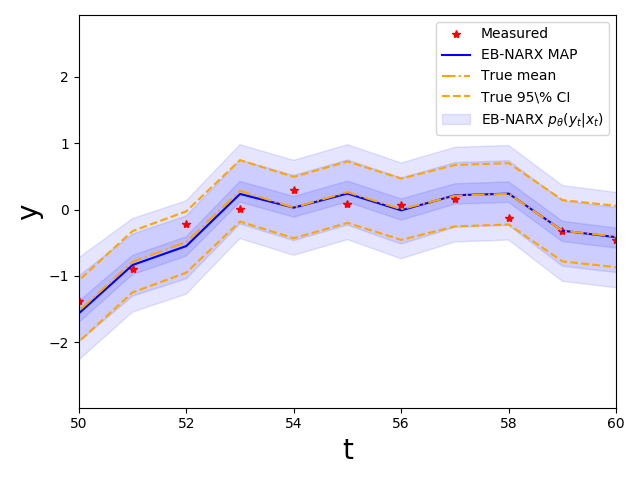}}
    \subcaptionbox{$t=56$ \label{subfig:arx_single}}{
    \includegraphics[width=0.3\linewidth]{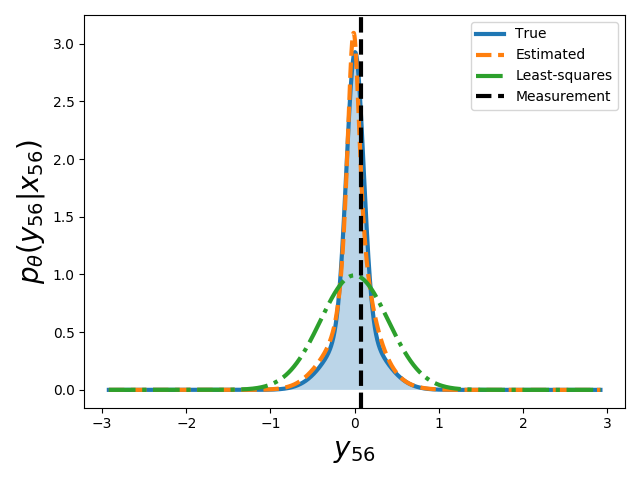}}

    \caption{(\subref{subfig:arx_sequence}) Estimates of
      $p_\theta(y_t | x_t)$ for a validation data sequence. The blue shading indicates the $65\%$, $95\%$, $99\%$ confidence regions. (\subref{subfig:arx_single}) The EB-NARX and least-squares estimates and true distribution for $t=56$.}
    \label{fig:arx}
\end{figure}

To further build confidence in the method's ability to learn the
distribution $p_\theta(y_t | x_t)$, it is demonstrated on data
generated using a second-order linear autoregressive eXogenous (ARX)
model;
\begin{equation}
    y_t = 1.5y_{t-1}-0.7y_{t-2}+u_{t-1}+0.5u_{t-2} + e_t,
\end{equation}
where $e_t\sim 0.6\mathcal{N}(0,0.1^2)+0.4\mathcal{N}(0,0.3^2)$.
An EB-NARX model is trained on 1000 data points and then used to predict the distribution for 200 validation data points. Figure~\ref{fig:arx}\subref{subfig:arx_sequence} shows part of the predicted sequence along with the true mean and $95\%$ confidence interval (CI). Figure~\ref{fig:arx}\subref{subfig:arx_single} shows the prediction $p_\theta(y_t | x_t)$ for $t=56$ given by the EB-NARX model and an ML estimate given by least-squares\footnote{This ML estimate makes an implicit Gaussian assumption.}, compared to the true Gaussian mixture distribution.
This illustrates that the EB-NARX model is able to accurately learn the mixture distribution and provide significantly more accurate quantification of the uncertainty than a standard ML approach. 

\subsection{Simulated Nonlinear Problem} 
\label{sub:nonlinear_arx}
\begin{figure}[!htb]
    \centering
    \subcaptionbox{Sequence}{
    \includegraphics[width=0.3\linewidth]{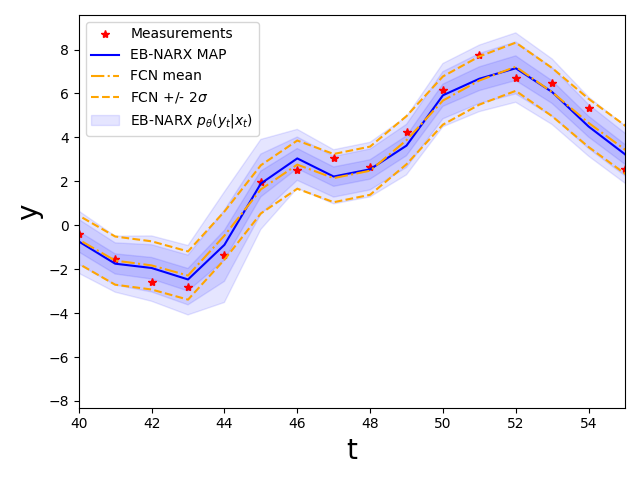}}
    \subcaptionbox{$t=53$ }{
    \includegraphics[width=0.3\linewidth]{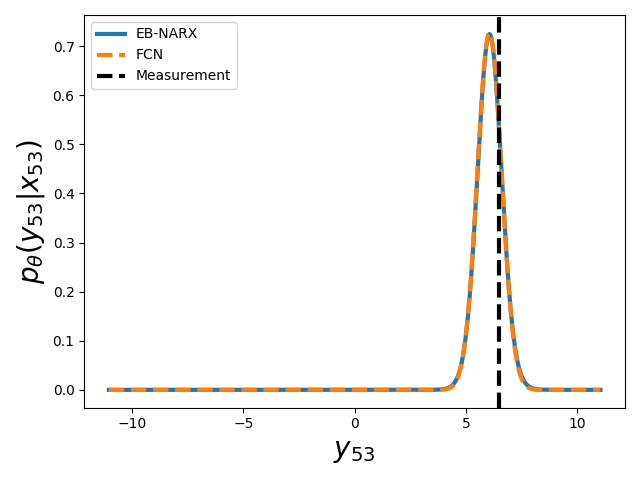}}

    \caption{Estimates of $p_\theta(y_t | x_t)$ for a validation data set generated using the nonlinear ARX model presented by \cite{chen1990non}. The blue shading indicates the $65\%$, $95\%$, $99\%$ confidence regions.}
    \label{fig:chen}
\end{figure}

So far, the method has been demonstrated on linear problems for which the learned distributions could be easily compared to the true distributions.
The method is now applied to data simulated using the nonlinear model \citep{chen1990non};
\begin{equation}\label{eq:chen_model}
\begin{split}
    y^*_t =& \left(0.8 - 0.5e^{-y^{*2}_{t-1}}\right)y^*_{t-1} - \left(0.3+0.9e^{-y^{*2}_{t-1}}\right)y^*_{t-2}\\ &+ u_{t-1}+0.2u_{t-2}+0.1u_{t-1}u_{t-2}+v_t,\\
    y_{t} =& y^*_t+w_t,
\end{split}
\end{equation}
where $v_t \sim \mathcal{N}(0,\sigma_v^2)$ and $w_t \sim \mathcal{N}(0,\sigma_w^2)$.
Using $D_u=D_y = 2$, the performance of the EB-NARX model is compared to that of an FCN for a range of noise standard deviations and training sequence lengths in Table~\ref{tab:chen_comparison}. 
These results indicate that the EB-NARX model performs competitively with the FCN for this data despite making no assumptions about the form of the distribution.
An example of the predicted distributions for data generated using $\sigma_v=\sigma_w=0.3$ and $N=1000$ is shown in Figure~\ref{fig:chen}. Since training the FCN using a squared-error loss function implicitly assumes Gaussian noise, it is, therefore, possible to determine the Gaussian distribution for the estimates and compare this to the distribution learned using the EB-NARX model. The variance of the FCN distribution has been calculated as the sample variance. 

\newcommand{\ra}[1]{\renewcommand{\arraystretch}{#1}}
\begin{table}[!htb]
\ra{1.0}
\centering
\captionsetup{width=1.0\linewidth}
\caption{Simulated nonlinear MSE on the validation set for the FCN and EB-NARX model trained on datasets generated with different noise levels ($\sigma_v = \sigma_w = \sigma$) and lengths (N). Only the best results are reported from among the different hyper-parameters and architectures considered --- the selection of which is detailed in Appendix~\ref{sec:hyper_parameter_and_structure_selection}.}
\label{tab:chen_comparison}
\scalebox{0.87}{\begin{tabular}{@{}l|rr|rr|rr|@{}}\toprule
& \multicolumn{2}{c}{$N=100$} & \multicolumn{2}{c}{$N=250$} & \multicolumn{2}{c}{$N=500$} \\
 & FCN & EB-NARX & FCN & EB-NARX & FCN & EB-NARX  \\
\midrule
$\sigma = 0.1$ & 0.122 & 0.099 & 0.069 & 0.070 & 0.057 & 0.054  \\
$\sigma = 0.3$ & 0.398 & 0.390 & 0.353 & 0.354 & 0.289 & 0.308  \\
$\sigma = 0.5$ & 0.860 & 0.869 & 0.809 & 0.822 & 0.754 & 0.779 \\
\bottomrule
\end{tabular}}
\end{table}

\subsection{Real Data: Coupled Electric Drives} 
\label{sub:real_data_coupled_electric_drives}
\begin{figure}[!htb]
    \centering
    \includegraphics[width=0.4\linewidth]{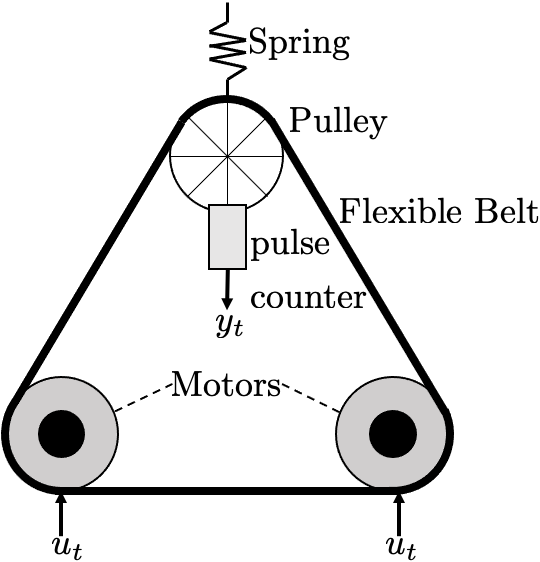}
    \caption{Illustration of the CE8 coupled electric drives system \citep{wigren2017coupled}.}
    \label{fig:CED_diagram}
\end{figure}

We now demonstrate the practical utility of the presented method by
application to the CE8 coupled electric drives benchmark data set
\citep{wigren2017coupled}. The coupled electric drives system, illustrated in
Figure~\ref{fig:CED_diagram}, consists of two electric motors that drive a pulley using a
flexible belt. The pulley is held by a spring and its angular speed is
measured by a pulse counter, which is insensitive to the sign of the
angular velocity. This creates an ambiguity in the measurements. The
input to the system is the signal sent to both motors.

The first three data sets described in \citet{wigren2017coupled},
which use a random binary input signal, were combined and split
$50/50$ between training and validation, giving 750 data points
each. This data was used to train an FCN and an EB-NARX model, with the
delays $D_u=D_y=3$ and the selection of hyperparameters and structure
detailed in
Appendix~\ref{sec:hyper_parameter_and_structure_selection}. 

\begin{figure}[!htb]
    \centering
    \subcaptionbox{$p_\theta(y_t | x_t)$ sequence}{
    \includegraphics[width=0.3\linewidth]{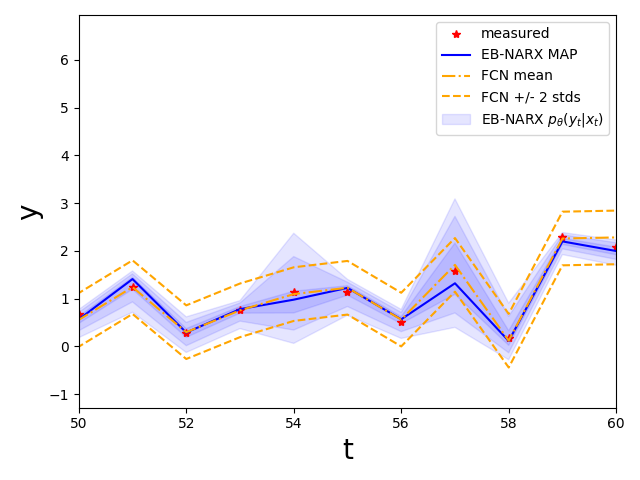}}
    \subcaptionbox{$t=40$ }{
    \includegraphics[width=0.3\linewidth]{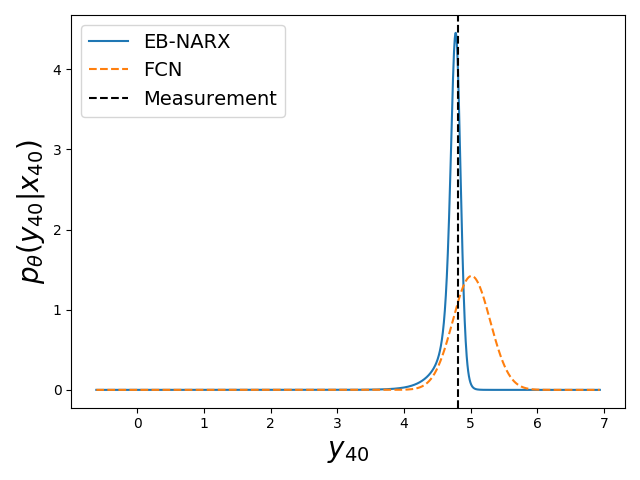}}\\
    \subcaptionbox{$t=57$ }{
    \includegraphics[width=0.3\linewidth]{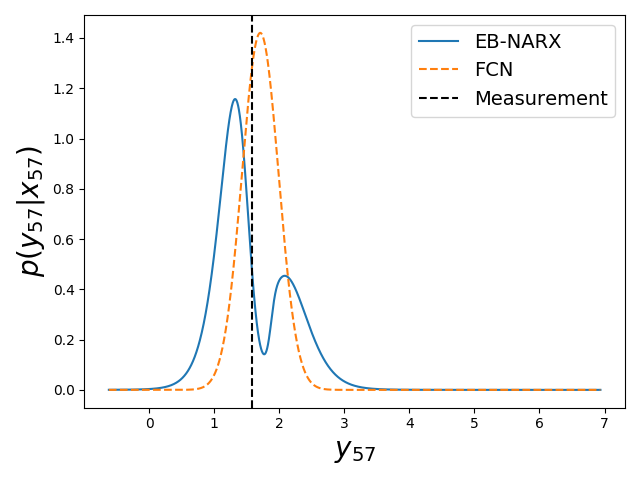}}
        \subcaptionbox{$t=60$ }{
    \includegraphics[width=0.3\linewidth]{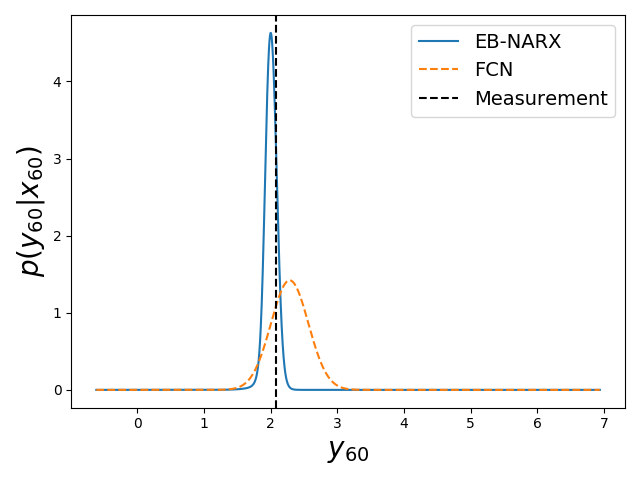}}

  \caption{Estimates of $p_\theta(y_t | x_t)$ for a sequence of
    validation data from the CE8 coupled electric drives benchmark data set
\citep{wigren2017coupled}. The blue shading indicates the $65\%$, $95\%$, $99\%$ confidence regions. The sample variance was used to determine the
    variance of the FCN assumed Gaussian distribution.}
    \label{fig:ced}
\end{figure}

The best result for the FCN was an MSE of $0.0521$, and for the EB-NARX model an MSE of $0.0503$. Figure~\ref{fig:ced} shows examples of
estimates produced using the FCN and EB-NARX models. As in Section~\ref{sub:nonlinear_arx}, the sample variance has been used for the Gaussian
distribution of the FCN prediction. This variance is constant for all
time steps, whereas the EB-NARX model predicts distinctly different
and even non-Gaussian distributions at some time steps.

This example demonstrates the flexibility of the EB-NARX model since the magnitude of the angular velocity is measured rather than the angular velocity itself. This produces a sign ambiguity, which has an impact when the velocity crosses zero (there is a reflection in the speed). Intuitively, we expect the measurement distribution to be multi-modal around these points and indeed this intuition is supported by the estimates from the EB-NARX model.
In contrast, the sample variance for the FCN predictions does not capture the dependence of the distribution on $x_t$ and therefore over-estimates the variance away from zero and under-estimates it close to zero.


\section{Conclusion \& Discussion} 
\label{sec:conclusion}

The salient feature of the EB-NARX model is that it has a highly flexible functional form, which is capable of adapting both to simple and more complex distributions. 
By contrast, more traditional approaches typically assume a noise distribution that is convenient for learning purposes. While the examples demonstrate that this flexibility is quite useful, it should be noted that the comparisons presented in this work only considered a relatively limited number of data sets, model types, and model structures. As such, a more thorough comparison should be undertaken as future work.

Given that the EB-NARX model is learning the full conditional distribution rather than the point estimate, it might be expected that the performance of the point predictions would suffer when compared to the standard application of an FCN. However, for the particular data sets studied in the nonlinear simulation example, the results in Table~\ref{tab:chen_comparison} indicate that the EB-NARX model approach gives competitive point estimates.
Further, when applied to a real data set from the CE8 coupled electric drives system, the EB-NARX model gave point estimates with a lower MSE than the estimates from a standard FCN. This suggests that the EB-NARX model may be a better choice when the conditional distribution depends on the current state of the system. 

In this work, the EB-NARX model was composed of two networks; a predictor net and a feature net. This structure is suggested by \cite{gustafsson2020energy} in the context of regression tasks with high dimensional input spaces, such as images. Hence, it may be less beneficial in the current setting where $x_t$ is typically of relatively low dimension.
The exploration of other structures that may be more suitable in the system identification context is another avenue for future research.

A limitation of the presented work is that it only considers one-step-ahead predictions and not multi-step-ahead predictions or even free-run simulations. Since the EB-NARX model predicts the full conditional distribution yet it takes as inputs point data, it is not clear how these predictions could be propagated forward in time. Whilst it would be possible to propagate the MAP estimate this does remove the main benefit over the standard FCN approach and further has questionable validity if the distribution is multimodal. 

Finally, the presented work has only considered NARX systems and an interesting area of future research would be to consider deep EBM's for other types of system identification problems.

\section*{Acknowledgements} 
\label{sec:acknowledgements}
This research was financially supported by the projects  \emph{Learning flexible models for nonlinear dynamics} (contract number: 2017-03807), \emph{NewLEADS -- New Directions in Learning Dynamical Systems} (contract number: 621-2016-06079), by the Swedish Research Council, by the Brazilian research agency CAPES and by \emph{Kjell och M{\"a}rta Beijer Foundation}.


\begin{appendix}
\section{Hyper-parameter and Structure Selection} 
\label{sec:hyper_parameter_and_structure_selection}
For each data set, $500$ FCN and EBM models were trained covering a range of structures and hyper-parameters. 
For the FCN, the number of layers ranged from 2 to 4. The dimension of each layer was varied from 50 to 300, and both \texttt{tanh} and \texttt{ReLU} activation functions were considered. 
For the EBM, the feature net was composed of two fully connected layers with \texttt{ReLU} nonlinearities and for the predictor net a neural network with four layers, \texttt{tanh} nonlinearities and skip connections. The hidden dimension of both the feature and predictor net was varied from 50 to 300.

For the training of both networks, batch sizes of 32, 64 and 128 were considered and training was carried out until the cost had plateaued. An initial learning rate of 0.001 with a decay rate of 0.99 was used in all cases. A different random seed was used to initialise the parameters each time.


\end{appendix}


\bibliography{Ref.bib}

\end{document}